\pdfoutput=1
\documentclass[journal]{IEEEtran}

\usepackage{cite}
\usepackage{float}

\ifCLASSINFOpdf
\usepackage[pdftex]{graphicx}

\usepackage{amsmath}
\usepackage{amssymb}
\usepackage{bm}
\usepackage{diagbox}

\DeclareMathOperator{\relu}{ReLU}

\DeclareMathOperator{\MLP}{MLP}

\DeclareMathOperator{\trace}{trace}
\newcommand\norm[1]{\left\lVert#1\right\rVert}

\usepackage{algorithmic}
\ifCLASSOPTIONcompsoc
\usepackage[caption=false,font=normalsize,labelfont=sf,textfont=sf]{subfig}
\else
\usepackage[caption=false,font=footnotesize]{subfig}

\usepackage{lipsum}
\usepackage{footnote}
\usepackage{hyperref}
\usepackage{stfloats}

\usepackage{url}
\usepackage{threeparttable}
\usepackage{multirow}

\usepackage{color}
\begin{document}
\bibliographystyle{IEEEtran}
\bstctlcite{IEEEexample:BSTcontrol}
%
\title{Ensemble manifold based regularized multi-modal graph convolutional network for cognitive ability prediction}
%
%
%

\author{Gang Qu,
        Li Xiao, 
        Wenxing Hu,
        Kun Zhang,
        Vince D. Calhoun,~\IEEEmembership{Fellow,~IEEE},
        Yu-Ping Wang,~\IEEEmembership{Senior Member,~IEEE}
\thanks{This work was supported in part by NIH under Grants R01 GM109068, R01 MH104680, R01 MH107354, P20 GM103472, R01 REB020407, R01 EB006841, and 2U54MD007595, and in part by NSF under Grant $\#$1539067. (Corresponding author: Yu-Ping Wang.) }
\thanks{Yu-Ping Wang* is with the Biomedical Engineering Department, Tulane University, New Orleans, LA 70118, USA. (e-mail: wyp@tulane.edu).}
\thanks{Kun Zhang is with the Department of Computer Science Department, Xavier University of Louisiana, New Orleans,LA 70125, USA. (e-mail: kzhang@xula.edu).
}
\thanks{Vince D. Calhoun is with the  Tri-Institutional Center for Translational Research in Neuroimaging and Data Science (TReNDS) (Georgia State University, Georgia Institute of Technology, Emory University), Atlanta, GA, USA (e-mail: vcalhoun@gsu.edu).}

\thanks{Gang Qu, Li Xiao, and Wenxing Hu are with the Biomedical Engineering Department, Tulane University, New Orleans, LA 70118, USA. (e-mail: gqu1@tulane.edu).}
}

\maketitle
\begin{abstract}
Objective: Multi-modal functional magnetic resonance imaging (fMRI) can be used to make predictions about individual behavioral and cognitive traits based on brain connectivity networks. Methods: To take advantage of complementary information from multi-modal fMRI, we propose an interpretable multi-modal graph convolutional network (MGCN) model, incorporating the fMRI time series and the functional connectivity (FC) between each pair of brain regions. Specifically, our model learns a graph embedding from individual brain networks derived from multi-modal data. A manifold-based regularization term is then enforced to consider the relationships of subjects both within and between modalities. Furthermore, we propose the gradient-weighted regression activation mapping (Grad-RAM) and the edge mask learning to interpret the model, which is used to identify significant cognition-related biomarkers. Results: We validate our MGCN model on the Philadelphia Neurodevelopmental Cohort to predict individual wide range achievement test (WRAT) score. Our model obtains superior predictive performance over GCN with a single modality and other competing approaches. The identified biomarkers are cross-validated from different approaches. Conclusion and Significance: This paper develops a new interpretable graph deep learning framework for cognitive ability prediction, with the potential to overcome the limitations of several current data-fusion models. The results demonstrate the power of MGCN in analyzing multi-modal fMRI and discovering significant biomarkers for human brain studies.
\end{abstract}
\IEEEpeerreviewmaketitle

\section{Introduction}
\label{Introduction}
Functional magnetic resonance imaging (fMRI) provides a non-invasive, high-resolution technique for observing the low-frequency fluctuation in blood-oxygenation-level-dependent (BOLD) signals to characterize the metabolism of the human brain. Recent evidence \cite{adeli2019multi, calhoun2016multimodal, cetin2014thalamus} suggests that multiple fMRI datasets contain complementary information and can predict individual variations in behavioral and cognitive traits better than using a single dataset. Numerous data fusion methods have been developed to integrate multiple paradigms of fMRI. For instance, ICA-based approaches \cite{calhoun2006method, sui2011discriminating} were proposed by Calhoun \textit{et al.} and Sui \textit{et al.} to analyze the joint information from multiple fMRI paradigms. Jie \textit{et al.} \cite{jie2015manifold} and Zhu \textit{et al.} \cite{zhu2017novel} proposed manifold regularized multi-task learning models to describe the subject-subject and the response-response relationships. These models were further extended by Xiao \textit{et al.} \cite{xiao2019manifold} to incorporate the relation information both within and between modalities. These approaches are typically based on linear models without considering complex nonlinear relationship between these data. 

Recently, there has been growing interest in applying graph deep learning models \cite{kipf2017semi} to various fMRI studies such as disease prediction \cite{parisot2018disease} and biomarkers identification \cite{li2019graph}. Compared with conventional deep learning models, graph deep learning model directly takes graph-structured data as the input and is therefore ideal for brain network analysis. To this end, we propose an interpretable end-to-end multi-modal graph convolution network (MGCN) framework to integrate multiple paradigms of fMRI, which incorporates different levels of information, from the signal in each brain region of interest (ROI) and the functional connectivity (FC) \cite{calhoun2014chronnectome} between each pair of ROIs to subject-subject and paradigm-paradigm relationships. The individual brain network is regarded as a graph represented using the ROIs as nodes and the FCs between each pair of ROIs as edges to predict the individual phenotypes. Specifically, the node feature on the graph is defined as the fMRI time-series of each ROI and the edge is defined as the FC between each pair of ROIs measured with Pearson correlation. We use the feature map learned from the intermediate layer of the network as the graph embedding to calculate individual similarities within and between paradigms. A manifold based regularization term is then enforced on the loss function to incorporate relationships between subjects. Our end-to-end MGCN model is then used for the phenotype prediction and biomarker identification. However, the decision mechanism behind the end-to-end neural network makes the model difficult to interpret. Inspired by the works of \cite{selvaraju2017grad, chattopadhay2018grad, wang2018diabetic}, we propose the gradient-based method, namely Gradient-weighted Regression Activation Mapping (Grad-RAM) to track each node's gradient and generate an activation map to identify significant nodes (ROIs) in the prediction task. Moreover, to interpret the model at the connection level, we propose the edge mask learning to learn a matrix pattern and show the significance of each edge (FC). The $L_{1}$ regularization is applied to analyze the pattern with different sparsity thresholds.

We validate our proposed framework using the Philadelphia Neurodevelopmental Cohort (PNC) \cite{satterthwaite2014neuroimaging,satterthwaite2016philadelphia}. Two fMRI paradigms, including a memory task and an emotion cognition task, referred as modalities here, are simultaneously used for phenotype prediction and biomarkers identification. The Wide Range Achievement Test (WRAT) \cite{test2009wide} score is used as the predicted phenotype, which measures individual ability in reading, spelling, comprehending, and solving mathematical problems. In the experiment, we first compare the predictive performance of our model with the other competing models. The results demonstrate that our proposed framework yields superior predictive performance. In addition, we interpret the framework and identify significant human cognition-related biomarkers in both ROI and FC levels using Grad-RAM and edge mask learning, respectively. The results from two interpretation strategies provide different views to understand the human brain's cognitive function. We then analyze the functional networks (FNs) based on the identification results and provide the relevant literature to support our findings.

We organize the rest of the paper as follows. In Section \ref{Methods}, we briefly review the graph convolutional networks, our proposed multi-modal graph neural network, and the details of Grad-RAM and edge mask learning. We then perform the experiment on PNC and describe the experimental results with some discussions in Section \ref{Experiments}. We conclude this paper in Section \ref{Conclusion}. We list the commonly used notations in Table \ref{Note} for convenience.
\begin{table}[ht!]
\renewcommand\arraystretch{1.2}
\begin{center}
    \caption{Commonly Used Notations.}
    \label{Note}
\begin{tabular}{|l|c|}\hline
      \textbf{Notation} & \textbf{Definition}\\
      \hline
      $\mathcal{G}$ & a graph \\

	$\norm{{\bm{B}}}_\mathrm{1}$ & the sum of the absolute values of all entries of matrix $\bm{B}$\\ 

	$\bm{B}_{(i,j)}$& the $(i, j)_{th}$ entry of matrix $\bm{B}$\\
	$\bm{B}^{(m)}_n$ & matrix $\bm{B}$ for the $n_{th}$ subject in the $m_{th}$ modality\\
	$H^l$& the feature map in the $l_{th}$ layer of the framework\\
	$N$ & number of subjects\\
	$Q$ & number of nodes (ROIs) in brain networks\\
	$C$ & number of feature channels for the graph embedding\\
    $\odot$ & the Hadamard product\\
	$\bm{X}$& the fMRI time-series\\
      \hline
    \end{tabular}
  \end{center}
\end{table}

\begin{figure*}[htbp!]
\centering
\includegraphics[width = 0.9\textwidth]{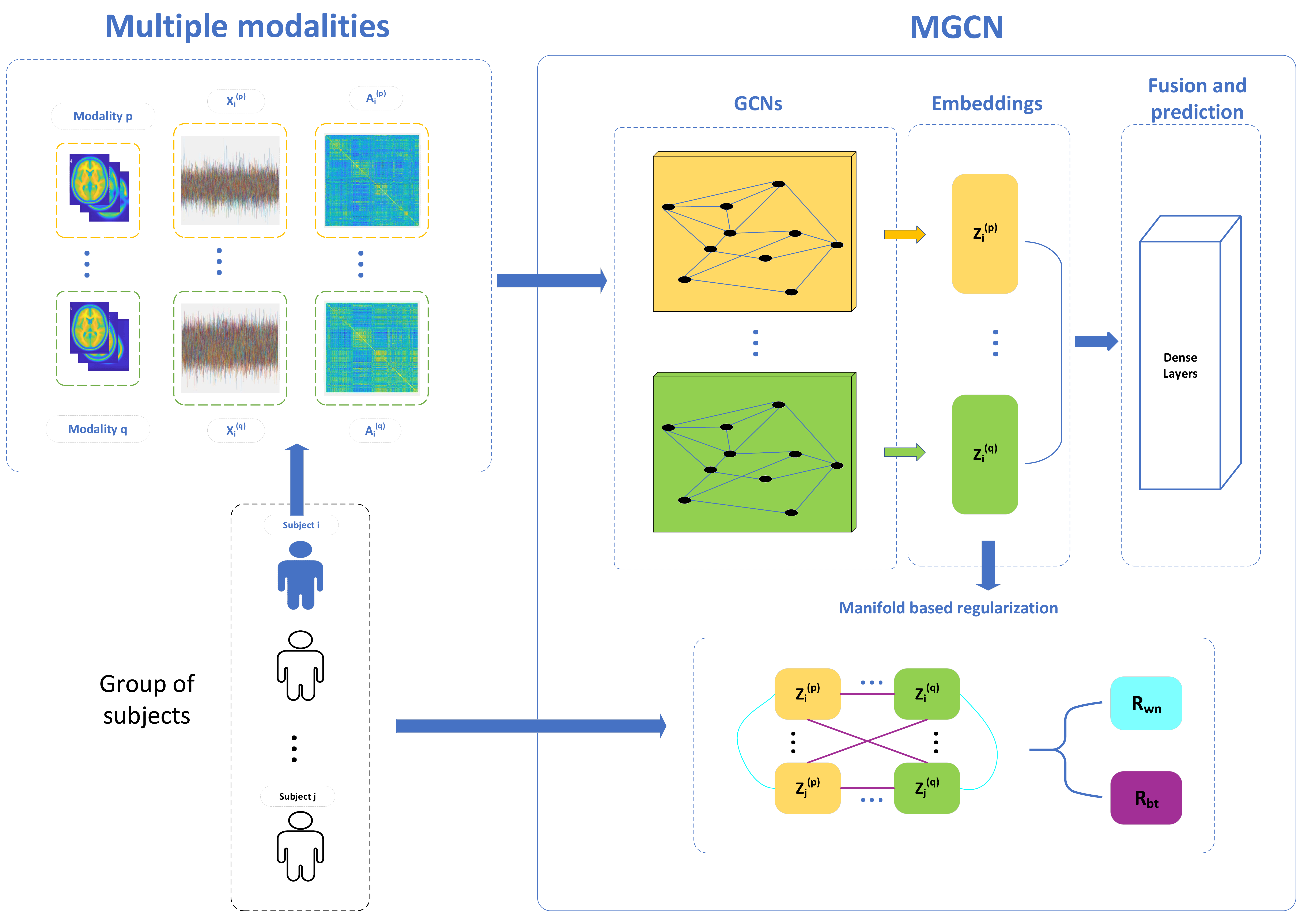}
\caption{The flowchart of our proposed framework. The GCN block can learn the graph embeddings from multi-modal. Afterward, the manifold based regularization is calculated to incorporate the relations of subjects within and between modalities. The dense layers fuse the embeddings from different modalities for prediction.}
\label{workflow}
\end{figure*}
 
\section{Methods}
\label{Methods}
\subsection{Graph convolutional network}
\label{Graph convolutional network}
Graph convolutional network (GCN) uses networks as the input. Given a graph $\mathcal{G}$, let $\bm{U}$ be the eigenvector matrix of the normalized graph Laplacian:
\begin{equation}
\bm{L}_{\mathrm{norm}}:=\bm{I}-\bm{D}^{-\frac{1}{2}}\bm{A}\bm{D}^{-\frac{1}{2}},
\label{norm lap}
\end{equation}
where $\bm{I}$, $\bm{D}$, and  $\bm{A}$ are the identity matrix,  the degree matrix, and the adjacency matrix, respectively. The graph convolution between the graph signal $\bm{x}$ and the filter $g_{\theta}$ in the spatial domain \cite{bruna2014spectral} can be defined as
\begin{equation}
g_{\theta}*\bm{x}=\bm{U}g_{\theta}\bm{U}^{\mathrm{T}}\bm{x},
\label{graph convolution}
\end{equation}
where $\bm{x}\in\mathbb{R}^{d}$ is the input vector for each node. 

To reduce computational complexity, graph convolution can be further simplified using Chebyshev polynomials up to the $K_{th}$ order into
\begin{equation}
g_{\theta}*\bm{x}=\sum^{K}_{k=0}\theta_k^{\prime}T_k(\tilde{\bm{L}}),
\label{cheby graph convolution}
\end{equation}
where $\tilde{\bm{L}}=2\bm{L}_{norm}/\lambda_{max}-\bm{I}$, and $\lambda_{max}$ refers to the largest eigenvalue of $\bm{L}_{norm}$, and $T_k$ is the $K_{th}$ order Chebyshev polynomial. The number of free parameters for graph convolution can be further constrained \cite{kipf2017semi} using renormalization. Therefore, the GCN layer can be rewritten as
\begin{equation}
\bm{H}^{l+1}= \phi^l(\tilde{\bm{D}}^{-\frac{1}{2}}\tilde{\bm{A}} \tilde{\bm{D}}^{-\frac{1}{2}}\bm{H}^l\Theta^l)\\
\label{simplify graph convolution}
\end{equation}
where $\tilde{\bm{A}}=\bm{A}+\bm{I}$, $\tilde{\bm{D}}$ is the degree matrix with respect to $\tilde{\bm{A}}$, and $\phi^l$, $\bm{H}^l$ and $\Theta^l$ are the activation function, the feature map and the weight matrix for the $l_{th}$ layer, respectively.

\subsection{Multi-modal graph convolutional networks}
\label{Proposed model}
We next propose our multi-modal graph convolutional networks (MGCN) to analyze the brain networks derived from multiple modalities. The brain network for each individual is regarded as one graph. The edge is here represented using absolute value of the Pearson correlation coefficient between the time-series on each pair of ROIs. The K-nearest neighbor algorithm is then applied to the adjacency matrix, keeping the largest $K$ values for each row and column and letting the rest be zero. We apply the two-layer GCN to learn the graph embedding for an individual in each modality, i.e.,
\begin{equation}
\bm{Z}= \phi^1(\tilde{\bm{D}}^{-\frac{1}{2}}\tilde{\bm{A}} \tilde{\bm{D}}^{-\frac{1}{2}} \phi^0 (\tilde{\bm{D}}^{-\frac{1}{2}}\tilde{\bm{A}} \tilde{\bm{D}}^{-\frac{1}{2}}\bm{X}\bm{\Theta}^0)\bm{\Theta}^1),
\label{GCN block}
\end{equation}
where $\bm{X}$ and $\bm{Z}$ are the fMRI time-series and the graph embedding, respectively. Notably, the individuals in the same modality share the same $\bm{\Theta}^0$ and $\bm{\Theta}^1$.

The embeddings of all individuals in the same modality are vectorized and formed into a matrix $\bm{Z}^{(m)}=[vec(\bm{Z}_{1}^{(m)}),vec(\bm{Z}_{2}^{(m)}),\cdots,vec(\bm{Z}_{N}^{(m)})]^\mathrm{T}\in\mathbb{R}^{N\times QC}$. Next, the multi-layer perceptron (MLP) of two dense layers is used to fuse the embeddings from all $M$ modalities for the prediction.
\begin{equation}
\hat{\bm{y}}=\MLP(\bm{Z}),
\label{MLP}
\end{equation}
where $\bm{Z}=[\bm{Z}^{(1)};\bm{Z}^{(2)};\cdots;\bm{Z}^{(M)}]\in\mathbb{R}^{MN\times QC}$, and $\hat{\bm{y}}\in\mathbb{R}^{N}$ is the prediction of WRAT scores for $N$ subjects.

However, the relationships of subjects from different modalities have not been considered yet. For a joint analysis of multiple modalities to boost learning performance, we impose a novel manifold based regularization term, enabling the MGCN to incorporate the relationships between the subjects both within and between multiple modalities. To do so, we first used the Pearson correlation to define the brain network similarity of two subjects in the $m_{th}$ modality as 
\begin{equation}
    \bm{S}_{(i,j)}^{(m)} = |corrcoef(vec(\bm{A}_i^{(m)}), vec(\bm{A}_j^{(m)}))|,
\end{equation}
where $\bm{S}_{(i,j)}^{(m)}$ is the $(i, j)_{th}$ entry of the matrix $\bm{S}^{(m)}\in\mathbb{R}^{N\times N}$, and here $\bm{A}_i^{(m)}$ and $\bm{A}_j^{(m)}$ are the adjacency matrices before applying the K-nearest neighbor algorithm. Then, the similarity of subjects between different modalities can be represented as the product of the similarity matrices in each modality, i.e. , $\bm{S}^{(p)}\bm{S}^{(q)}$ is the similarity between $p_{th}$ and $q_{th}$ modalities. Our manifold based regularization term is therefore defined as a combination of
\begin{align}
R_{wn}&\!=\!\frac{1}{2}\sum^M_{m}\sum^{N}_{i,j}\bm{S}^{(m)}_{(i,j)}\lVert vec(\bm{Z}^{(m)}_{i})-vec(\bm{Z}^{(m)}_{j})\rVert_2^2, \\\nonumber
R_{bt}&\!=\!\frac{1}{2}\sum^M_{p\neq q}\sum^{N}_{i,j}(\bm{S}^{(p)}\bm{S}^{(q)})_{(i,j)}\lVert vec(\bm{Z}^{(p)}_{i})-vec(\bm{Z}^{(q)}_{j})\rVert_2^2,
\label{manifold multi}
\end{align}
where $R_{bt}$ and $R_{wn}$ are used to incorporate the subject-subject relationship within each single modality and between modalities, respectively. 

The manifold regularization term fully explores the relationship between subjects, enforcing the model to learn similar embeddings for subjects with high brain structure similarity both within and between modalities. Furthermore, we can write the similarities into a single matrix $\bm{S}\in\mathbb{R}^{MN\times MN}$ with the regularization parameters, 
\begin{equation}
\bm{S}=\left[ \begin{array}{llll}
\eta_2\bm{S}^{(1)}   & \eta_1\bm{S}^{(1)}\bm{S}^{(2)}   & \cdots   &\eta_1\bm{S}^{(1)}\bm{S}^{(M)}\\
\eta_1\bm{S}^{(2)}\bm{S}^{(1)}   & \eta_2\bm{S}^{(2)}   & \cdots   &\eta_1\bm{S}^{(2)}\bm{S}^{(M)}\\
\vdots                              & \vdots                               & \ddots   & \vdots                           \\
\eta_1\bm{S}^{(M)}\bm{S}^{(1)}  & \eta_1\bm{S}^{(M)}\bm{S}^{(2)}   & \cdots   &\eta_2\bm{S}^{(M)}
\end{array} 
\right ],\\
\label{big matrix}
\end{equation}
where the diagonal elements represent the within modality similarities, and $\eta_1$ and $\eta_2$ are the regularization parameters of between and within modalities, respectively. Therefore, the regularization term can then be written as
\begin{equation}
R_{m} = \eta_1R_{bt}+\eta_2R_{wn}= \trace(\bm{Z}^{\mathrm{T}}\bm{L}\bm{Z}),  
\label{my regularizer}
\end{equation}
where $\bm{L}=\bm{D}-\bm{S}$, and $\bm{D}$ is the degree matrix of $\bm{S}$. The mean square error (MSE) combined with the manifold based regularization term and the $L_{2}$ norm on weight matrices is used as the loss function (denoted as $loss$). The workflow of our MGCN framework is shown in Fig.\ref{workflow}.

\subsection{Gradient-weighted Regression activation mapping}
To interpret the MGCN at the node level, we propose the Gradient-weighted Regression Activation Mapping (Grad-RAM) to track the gradient of the graph embedding with respect to the prediction value in each modality,
\begin{equation}
\bm{G}^{(m)}_{(n,qc)}= \frac{\partial\bm{y}_n}{\partial\bm{Z}^{(m)}_{(n,qc)}},
\label{gradients}
\end{equation}
where $\bm{y}_n$ is the label for the $n_{th}$ subject, $\bm{G}^{(m)}\in\mathbb{R}^{N\times QC}$ is the gradient matrix in $m_{th}$ modality, and $\bm{Z}^{(m)}_{(n,qc)}$ is the $(n,qc)_{th}$ entry of $\bm{Z}^{(m)}$. 

Accordingly, using the graph embedding values to be the weights, we defined the Grad-RAM as the product between gradients and the graph embeddings over $N$ subjects.
\begin{equation}
\bm{a}^{(m)}_{(q)} = \frac{1}{NC}\relu(\sum_{n=1,c=1}^{N C}\bm{G}^{(m)}_{(n,qc)}\bm{Z}^{(m)}_{(n,qc)}),
\label{RAM}
\end{equation}
where $\bm{a}^{(m)}_{(q)}$ is the Grad-RAM value of the $q_{th}$ ROI in the $m_{th}$ modality. The ReLU function is applied to keep the features with a positive influence on the final prediction. Afterward, we have $\bm{a}^{(m)}\in\mathbb{R}^{Q}$ as the activation map in the ROI level for the $m_{th}$ modality. We used Grad-RAM to visualize the important ROIs for WRAT score prediction.

\subsection{Edge mask learning}
We next propose the edge mask learning to interpret our framework at the edge level. In particular, rather than learning explainable patterns for each modality separately, we retrain the model and jointly learn an edge mask matrix for subjects from all modalities. Since only the indirect graph is considered, the edge mask must be symmetric and non-negative, which is defined as
\begin{equation}
\bm{\mathcal{M}}=\relu(\bm{V}+\bm{V}^{\mathrm{T}}),
\end{equation}
where the entry of mask $\bm{M}$ indicates the weight for each edge, $\bm{V}\in\mathbb{R}^{Q\times Q}$ is the matrix we need to optimize, and the $\relu$ function ensures the elements to be non-negative. 

Consequently, the Eq.\ref{simplify graph convolution} and the loss function can be rewritten as Eq.\ref{mask GCN} and Eq.\ref{mask loss}, respectively.
\begin{equation}
    \bm{H}^{l+1}= \phi^l((\bm{\mathcal{M}}+\bm{I})\odot(\tilde{\bm{D}}^{-\frac{1}{2}}\tilde{\bm{A}} \tilde{\bm{D}}^{-\frac{1}{2}})\bm{H}^l\Theta^l),
    \label{mask GCN}
\end{equation}
\begin{equation}
    \hat{loss}=loss+\beta\norm{{\bm{\mathcal{M}}}}_\mathrm{1},
    \label{mask loss}
\end{equation}
where $\beta$ is the $L_{1}$ regularization parameter to control the sparsity level of the learned mask. Notably, by adding the identity matrix $\bm{I}$ to $\bm{\mathcal{M}}$, we obtain the identity mapping when $\bm{\mathcal{M}}=\bm{0}$ and guarantee the graph filter will not degrade to the null matrix.
\section{Experiments}
\label{Experiments}
Our framework was validated on the PNC. In this section, we first briefly describe the PNC and the preprocessing of the dataset, and then present our experimental results and discussions.

\subsection{Datasets}
\label{Datasets}
The PNC consists of multiple fMRI paradigms acquired from over 800 healthy subjects aged from 8 to 22 years. Two task fMRI data, the emotion cognition task (emoid-fMRI) and working memory task data (nback-fMRI), were used to predict various physiological phenotype. The emoid-fMRI and nback fMRI scan durations were 10.5 minutes (210 TR) and 11.6 minutes (231 TR). When measuring the emoid-fMRI, subjects were asked to identify the face with different emotions like angry, sad, fearful, happy, and label the emotion type. For the nback-fMRI scans, subjects were asked to conduct standard n-back tasks, which were related to working memory and the ability of lexical processing. The WRAT scores were collected from a 1-hour computerized neurocognitive battery (CNB) administered by the PNC,  which were regarded as the label to assess individual reading and comprehension ability \cite{wilkinson2006wide}.

\begin{table*}[hbt!]
\renewcommand\arraystretch{1.2}
  \begin{center}
    \caption{The prediction performance with different models.}
    \label{compare other models}
\begin{threeparttable}
    \begin{tabular}{|l|c|c|c|c|c|} 
    \hline
      \textbf{Model} & \textbf{Modalities}&\textbf{RMSE(mean $\pm$ std) }&\textbf{p-value}&\textbf{MAE(mean $\pm$ std)}&\textbf{p-value}\\
      \hline
GCN         & emoid fMRI & 15.6924 $\pm$ 0.9912    &    6.222e-04        &  12.3965 $\pm$ 0.7972 &  1.847e-04 \\ 
GCN         & nback fMRI  & 15.4335 $\pm$ 0.5867   & 9.022e-04  &12.0440 $\pm$ 0.6688 & 2.376e-04 \\ 
\hline
MTL         & emoid fMRI &   16.1742 $\pm$ 0.9283   &    2.548e-05  &  12.5691 $\pm$ 0.6846  & 2.101e-05  \\
MTL         & nback fMRI  &  16.2187 $\pm$ 0.9576   & 2.501e-05   &  12.5780 $\pm$ 0.7190  & 2.714e-05  \\
\hline
M2TL         & emoid fMRI &   15.9967 $\pm$ 0.8947   &    5.461e-05  &  12.5425 $\pm$ 0.5454  & 6.681e-06  \\
M2TL         & nback fMRI  &  15.0807 $\pm$ 0.8902   & 0.0136   &  11.6161 $\pm$ 0.7958  & 0.0362   \\
\hline
NM2TL         & emoid fMRI &   15.0320 $\pm$ 0.7965   &    0.0125  &  11.6963 $\pm$ 0.5853  & 0.0083  \\
NM2TL         & nback fMRI  &  15.0310 $\pm$ 0.7955   & 0.0125   &  11.6885 $\pm$ 0.5771  & 0.0085   \\
\hline
MLP  & emoid fMRI $\&$ nback fMRI &
15.1341 $\pm$ 0.5883 & 0.0023 & 11.9957 $\pm$ 0.9259 &0.0059\\
\hline
MVGCN & emoid fMRI $\&$ nback fMRI  &
14.9858 $\pm$ 0.9031& 0.0246 & 11.7140 $\pm$ 0.7608&0.0167\\
\hline
MGCN$^{\star}$ & emoid fMRI $\&$ nback fMRI &
14.8267 $\pm$ 0.6618 &0.0285 & 11.5339 $\pm$ 0.6226 &0.0337\\
MGCN & emoid fMRI $\&$ nback fMRI  & \textbf{14.0889} $\pm$ \textbf{0.7222}& -  &\textbf{10.8878} $\pm$ \textbf{0.6339}& - \\
\hline
    \end{tabular}
 \begin{tablenotes}
        \footnotesize
        \item[1] p-values were calculated by t-test between the regression performance for repeated experiments of our MGCN model and other competing models; std denotes the standard deviation.
        \item[2] $^\star$ the MGCN without the manifold based regularization term.
      \end{tablenotes}
    \end{threeparttable}
  \end{center}
\end{table*}

\begin{figure}[h!]\centering
\includegraphics[width =0.45\textwidth]{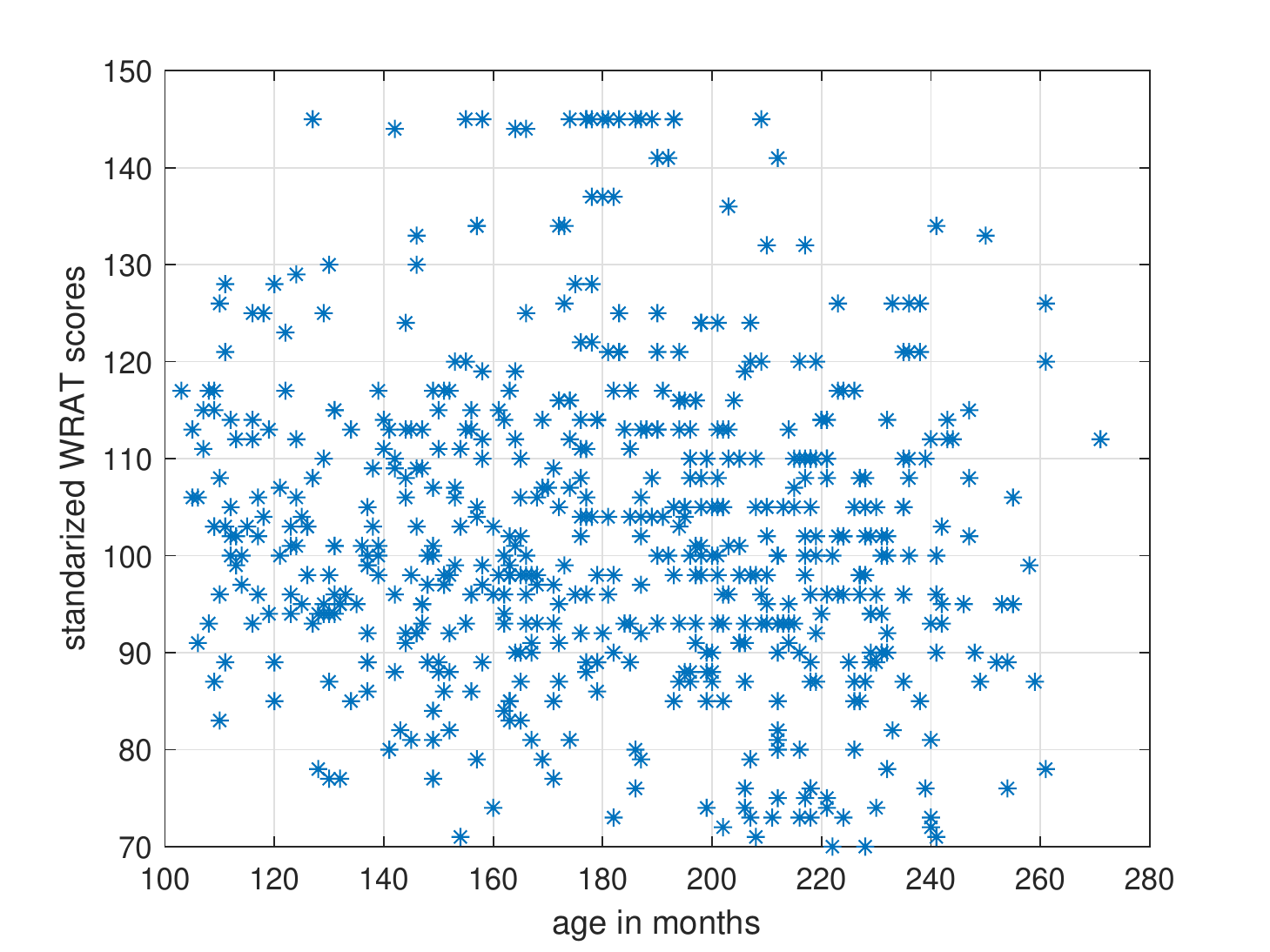}
\caption{The scatter plot of WRAT score versus age}\label{age_gender}
\end{figure}

All MRI scans were collected on a single 3T Siemens TIM Trio whole-body scanner. We followed similar preprocessing procedures as in \cite{zille2017enforcing, fang2017fast}. SPM12 \footnote{\url{http://www.fil.ion.ucl.ac.uk/spm/software/spm12/}} was used to conduct motion correction, spatial normalization, after which the data were smoothed with a 3mm Gaussian kernel. Multiple regressions were used while the influence of head motion was considered \cite{friston1995characterizing}. Then, we applied 264-region parcellation ($\bm{Q}=264$) to investigate whole-brain connectivity. As a result, we had 264 ROIs (containing 21,384 voxels) based on the Power coordinates template (sphere radius is 5mm) \cite{power2011functional}. We included only subjects with both emoid-fMRI and nback-fMRI resulting in 595 samples. We further investigated potential confounders for the WRAT-fMRI association, including age, gender, and head motion.
\begin{table}[h!]
\renewcommand\arraystretch{1.2}
\centering
\caption{Gender-WRAT information.}
\label{Gender}
\begin{tabular}{|l|c|c|c|}
\hline
property\textbackslash{}group & male   & female & Total  \\ \hline
number of subjects            & 271    & 324    & 595    \\ \hline
mean WRAT scores              & 104.20 & 101.44 & 102.30 \\ \hline
std WRAT scores               & 16.61  & 15.14  & 15.87  \\ \hline
\end{tabular}
\end{table}

\begin{itemize}
    \item Age: The WRAT-age distribution was shown in Table \ref{Gender}. Instead of using the raw WRAT scores, the standardized WRAT scores \cite{wilkinson2006wide} were considered, e.g., in Fig.\ref{age_gender}. The grade-based and age-based norms were added to control the age factor and to enhance the interpretation of the test results.
    \item Gender: The gender-WRAT information was shown in Table \ref{Gender}. The mean and standard deviation of WRAT scores of males and females are close. As a result, we consider the gender impact on WRAT scores is negligible..
    \item Head motion: We collected 6 rigid body motion parameters(3 translations along x, y, z axes, and 3 rotations around x, y, z axes). The framewise displacement (FD) of head position, which is defined as the sum of the absolute values of the derivatives of those 6 rigid body motion parameters \cite{power2014methods}, was then calculated. Next, we tested the hypothesis that there is no relationship between the WRAT scores and the mean FD using the matrix of Pearson correlation coefficients. The respective p-values were $0.3029$ for emoid-fMRI and $0.3599$ for nback-fMRI. No significant relationships between the head motion and WRAT scores were observed at $0.05$ level.
\end{itemize}

In addition, the mean-centering was applied to the labels using the mean WRAT scores of training samples. For each subject, the signal matrices were then formatted as $\bm{X}_{emoid}\in\mathbb{R}^{264\times 210}$ of the emoid-fMRI and $\bm{X}_{nback}\in\mathbb{R}^{264\times 231}$ of the nback-fMRI.
\begin{figure}[h!]\centering
\includegraphics[width = 0.45\textwidth]{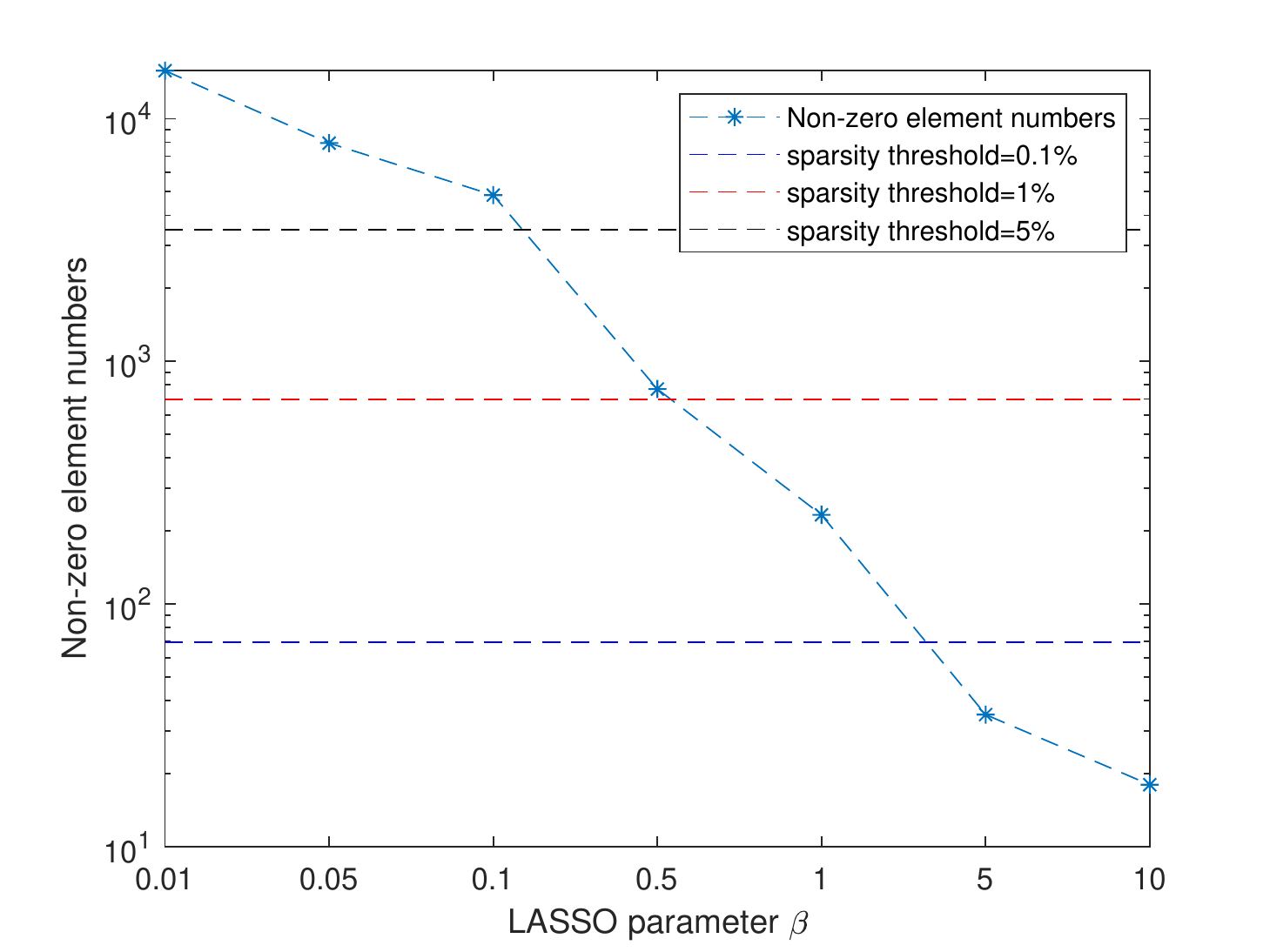}
\caption{The sparsity of edge mask with respect to the $L_{1}$ regularization parameter}\label{mask thre}
\end{figure}
\subsection{Experimental setup}
\label{Experimental setup}
We randomly split the data into training, validation, and testing sets with the ratio of $70\%$, $10\%$, $20\%$, respectively. We trained the model on the training set and tuned the hyperparameters on the validation set. The root mean square error (RMSE) and mean absolute error (MAE) between the predicted and observed WRAT scores in the test set were calculated. The bootstrap analysis was utilized to measure and compare models' performance, reducing sampling bias with 10 repeated experiments. For each experiment, we repeated the dataset splitting, model training, and testing processes. Besides, we compared the results of repeated experiments collected from our proposed model with other approaches, and reported the p-values of the pairwise t-test to show the statistically significant improvement.
\begin{table}[htb]
\renewcommand\arraystretch{1.2}
  \begin{center}
    \caption{The selected ROIs (top $5\%$)}
    \label{ROIs}
\begin{threeparttable}
    \begin{tabular}{|l|c|c|c|}
    \hline
	 \multirow{2}*{\textbf{ROI}} & \multicolumn{3}{c|}{\textbf{emoid-fMRI}} \\
	\cline{2-4}
	{}&\textbf{MNI space \tnote{1}}&\textbf{Anatomical region \tnote{1}}&\textbf{FN}\\
	\hline
	23&-23 -30 72& Postcentral L & SMN\\

	32&22 -42 69& Postcentral R & SMN\\

	47\tnote{*} & -3 2 53 & Supp Motor Area L & CNG\\

	59\tnote{*} & -5 18 34 & Cingulum Mid L & CNG\\

	92 & 8 -48 31 & Cingulum Mid R & DMN\\
	
	101 & 22 39 39 & Frontal Sup R & DMN\\

	112\tnote{*} & -2 38 36 & Frontal Sup R & DMN\\

	113\tnote{*} & -3 42 16 & Cingulum Ant L & DMN\\

	153\tnote{*} & 43 -78 -12 & Occipital Inf R & VIS\\

	162\tnote{*} & 24 -87 24 & Occipital Sup R & VIS\\
	
	167\tnote{*} & -3 -81 21 & Cuneus L & VIS\\

	213\tnote{*} & -1 15 44 & Supp Motor Area L & SAL\\

	215\tnote{*} & 0 -30 27& Cingulum Ant L & SAL\\
    \hline
	\textbf{ROI}&\multicolumn{3}{c|}{\textbf{nback-fMRI}}\\
	\hline
	47\tnote{*} & -3 2 53 & Supp Motor Area L & CNG\\

	57 & -34 3 4 & Clausrum & CNG\\

	59\tnote{*} & -5 18 34 & Cingulum Mid L & CNG\\

	112\tnote{*} & -2 38 36 & Frontal Sup Medial L & DMN\\

	113\tnote{*}& -3 42 16 & Cingulum Ant L & DMN\\
	
	145 & 8 -72 11 & Cakcarine R & VIS\\

	153\tnote{*} & 43 -78 -12 & Occipital Inf R & VIS\\

	162\tnote{*} & 24 -87 24 & Occipital Sup R & VIS\\

	167\tnote{*} & -3 -81 21 & Cuneus L & VIS\\

	202 & -3 26 44 & Frontal Sup Medial L & FPN\\
	
	213\tnote{*} & -1 15 44 & Supp Motor Area L & SAL\\

	215\tnote{*} & 0 30 27 & Cingulum Ant L & SAL\\

	229 & 31 -14 2 & Putamen R & SCT\\
	\hline
    \end{tabular}
 \begin{tablenotes}
        \footnotesize
        \item[*] indicates the ROIs common to emoid-fMRI and nback-fMRI
        \item[1] the coordinates of the ROIs' center point in Montreal Neurological Institute space and the anatomical regions \cite{schmahmann1999three} that the ROI locates in.
      \end{tablenotes}
    \end{threeparttable}
  \end{center}
\end{table}

\begin{figure*}[hbt!]
    \centering
    \includegraphics[width = \textwidth]{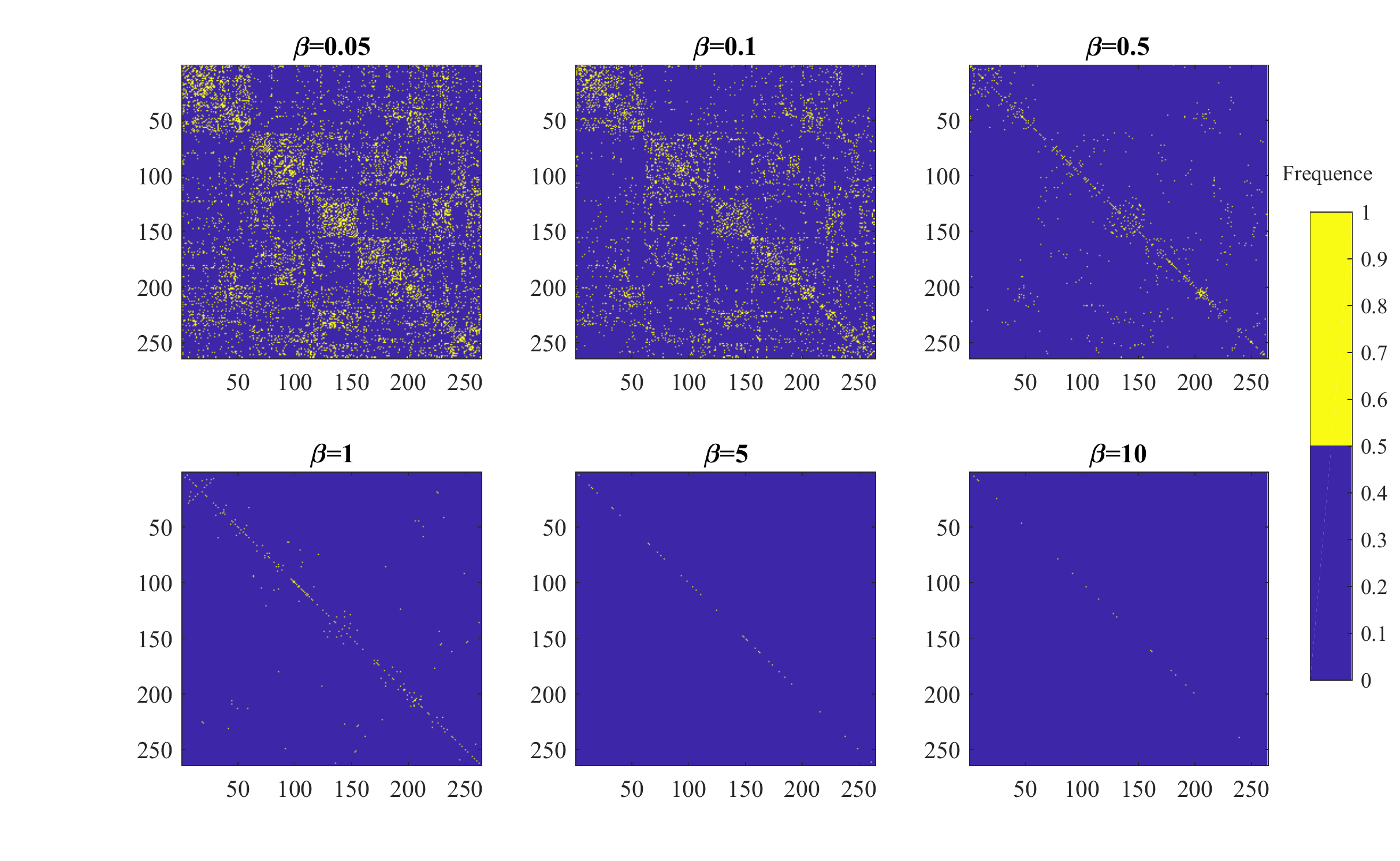}
    \caption{The edge masks learned with different $L_{1}$ regularization parameters}
    \label{mask plot}
\end{figure*}
\begin{figure}[hbtp!]
\centering
\includegraphics[width = 0.4\textwidth]{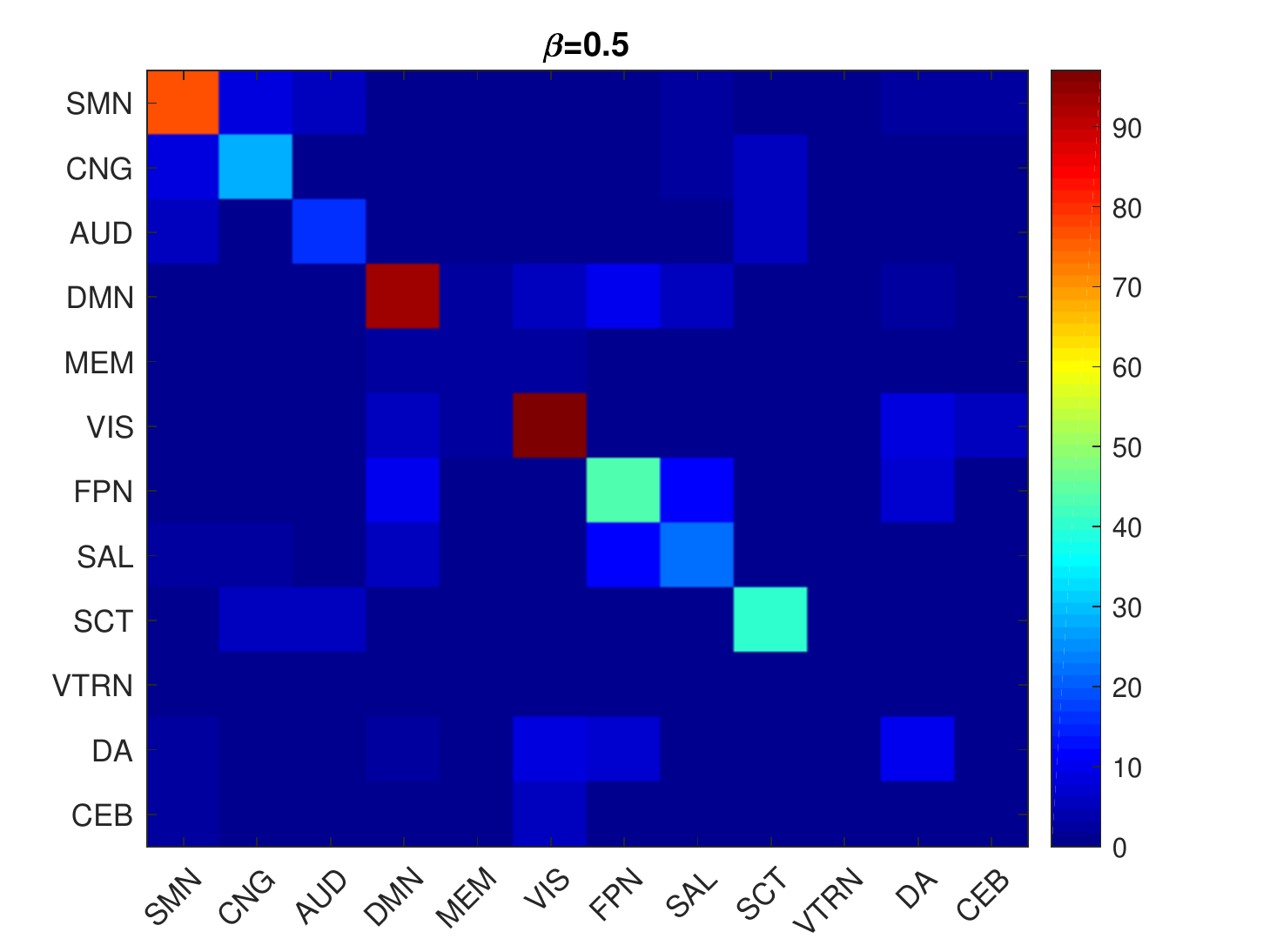}\\
\includegraphics[width =0.4\textwidth]{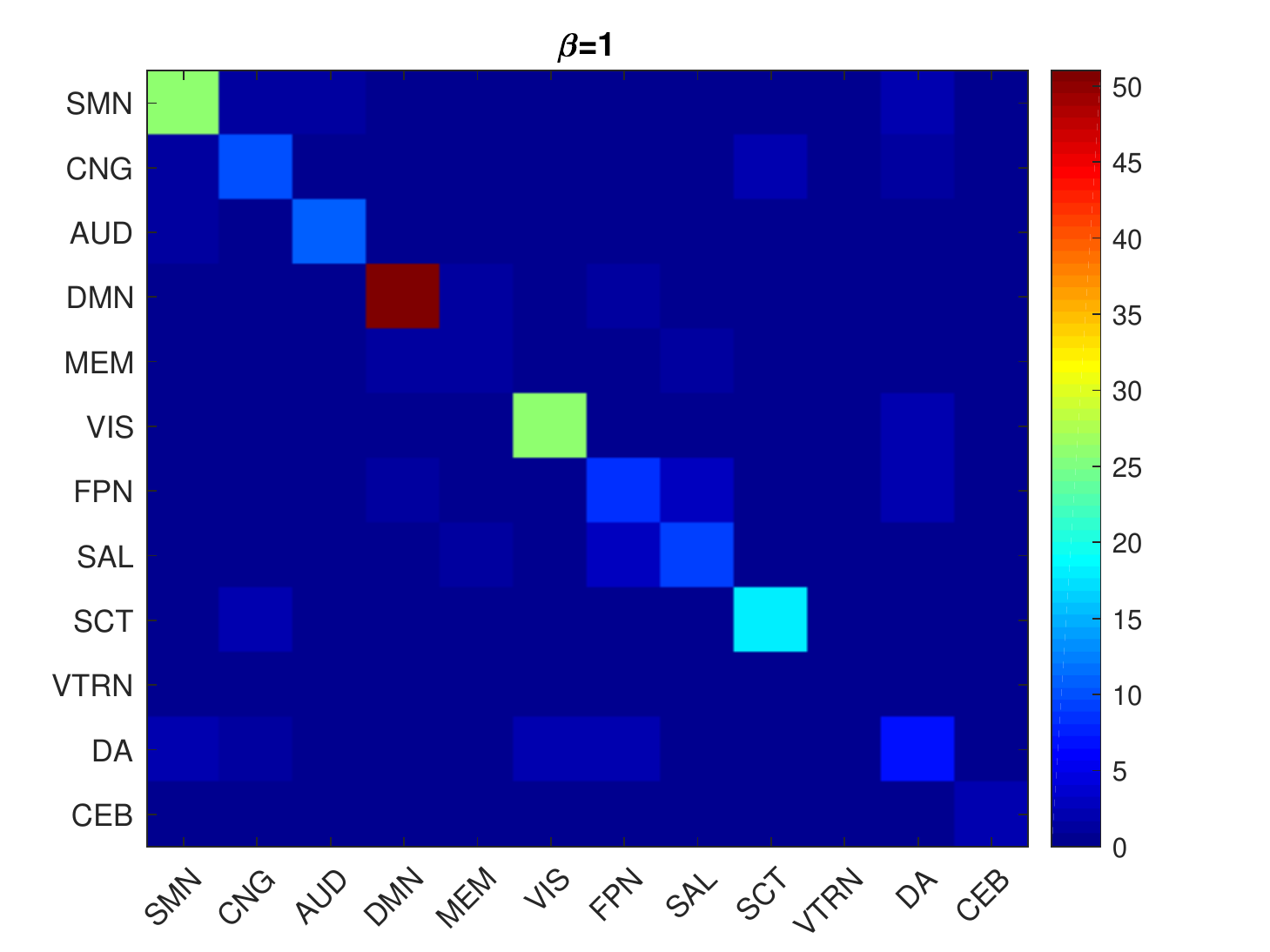}\\
\includegraphics[width =0.4\textwidth]{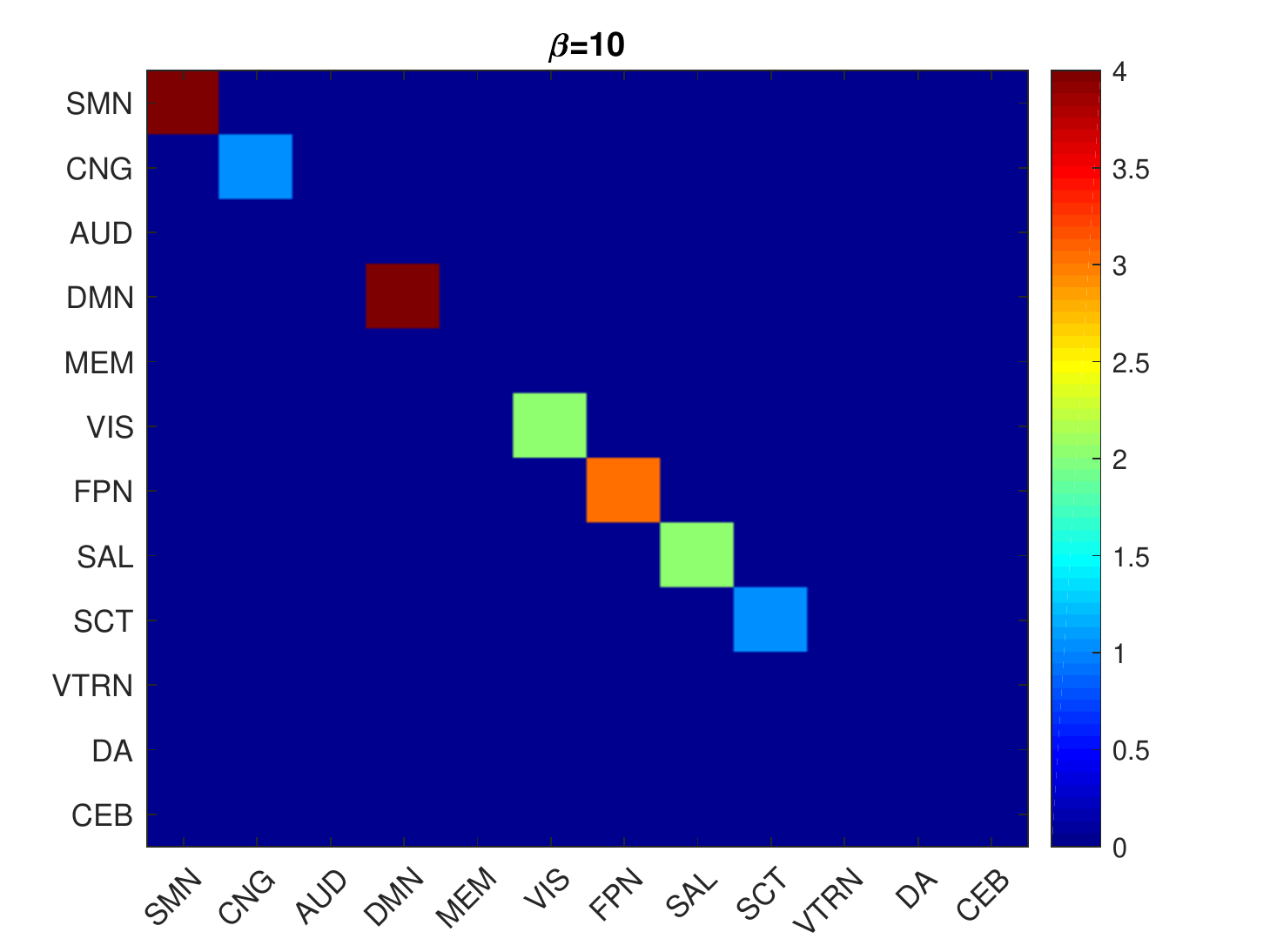}
\caption{Module allegiance matrices with different $L_{1}$ regularization parameters}
\label{FN map}
\end{figure}

\subsection{Results}
\label{Results}

\subsubsection{Hyperparameter selection}
The hyperparameters of the MGCN model were tuned on validation sets through the random search \cite{bergstra2012random}. In addition, the $L_{2}$ regularization term with a parameter $1e-4$ was enforced on weight matrices to avoid overfitting. Consequently, for MGCN, we set the $K=10$ for K-nearest neighbors and trained the model for up to 1000 epochs and a learning rate of $1e-5$ with early stop. The two-layer GCNs in Eq.\ref{GCN block} were used for graph embedding. We set the numbers of feature channels to be $128$ and $32$, and the activation functions to be $ReLu$ and $Sigmoid$ for the first and the second GCN layer, respectively. Next, two dense layers with respective $1024$ and $2048$ neurons and with $ReLU$ activation functions were used to merge the graph embeddings from two modalities, as in Eq.\ref{MLP}. The manifold based regularization parameters in Eq.\ref{my regularizer} were tuned in the range from $10^{-5}$ to $10^{-2}$. 

We here reported the running time in the training process and used GCN with the same graph convolution layers for a single modality as the baseline to measure the extra computational cost. The results showed that MGCN ($0.2132 \pm 0.0152 \textbf{ secs}$) took almost three times more time than GCN ($0.0796 \pm 0.0049\textbf{ secs}$) per iteration due to more complicated network structure and the calculation of manifold based regularization. 

\begin{figure*}[hbt!]
\centering
\includegraphics[width = 0.9\textwidth]{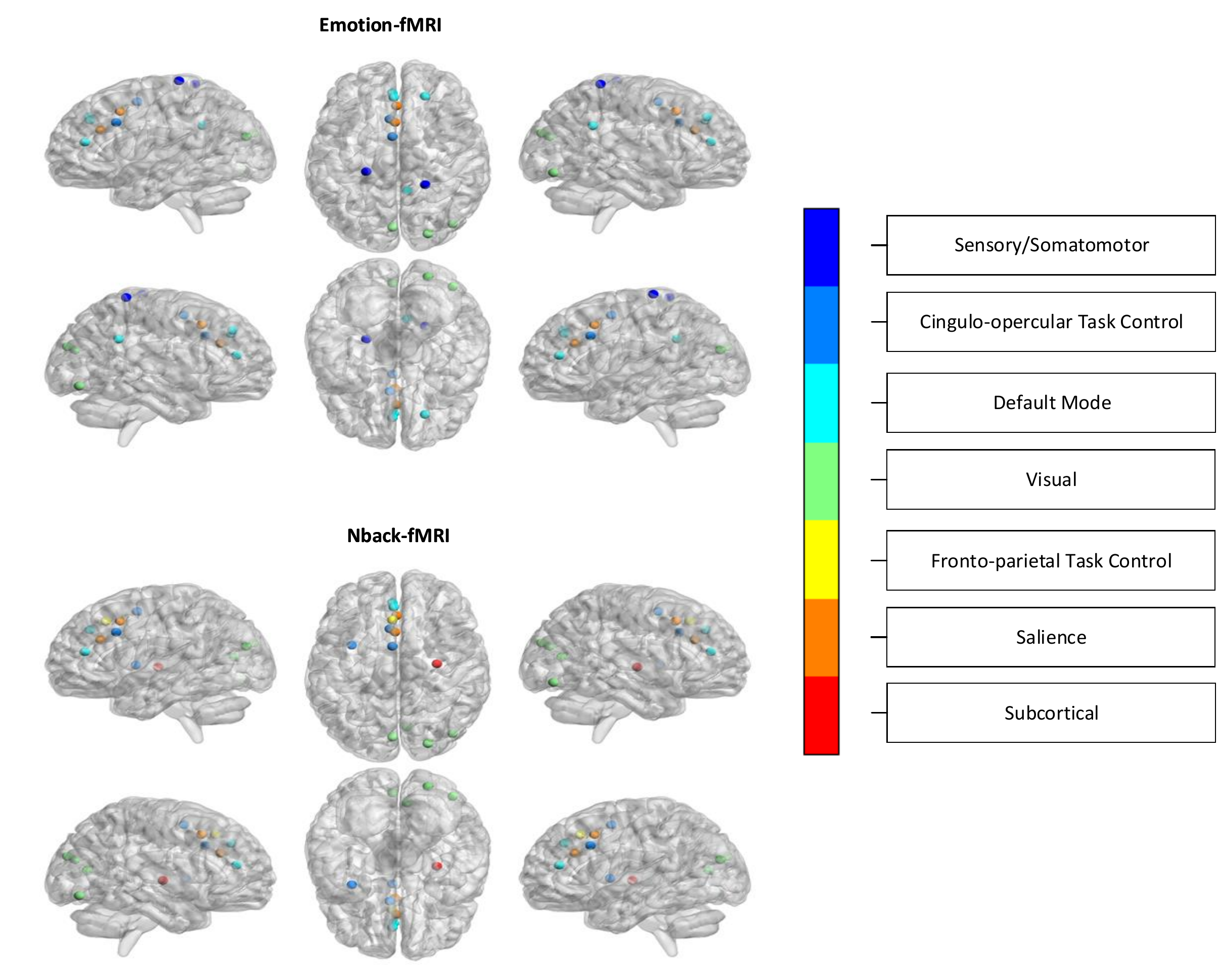}
\caption{The figure shows the anatomical view of selected ROIs ($5\%$) based on the power264 template. The same color indicates that the ROIs are in the same functional network. The upper parts are the results for emoid-fMRI; the lower parts are the results for nback-fMRI. Each figure contains the views from different layouts (sagittal, axial, coronal).}
\label{anatomyview}
\end{figure*}

\subsubsection{Performance in comparison to other methods}

We compared our model with the Multi-task learning model\cite{argyriou2006multi}, Manifold regularized multi-task learning (M2TL) \cite{zhu2017novel}, New Manifold regularized multi-task learning (NM2TL) \cite{xiao2019manifold}, GCN with single paradigm fMRI, MLP, multi-view GCN (MVGCN) \cite{zhang2018multi}, and MGCN without the manifold based regularization term in Eq.\ref{my regularizer}. 
\begin{itemize}
    \item The single modal GCN was validated using emoid-fMRI or nback-fMRI with the same hyperparameter selection as the GCN part of MGCN.
    \item For the MTL, M2TL, NM2TL, the hyperparameters were tuned using the random search. Notably, a different strategy in subjects selection was applied in \cite{xiao2019manifold}, in which the subjects were selected based on the age threshold for MTL, M2TL, and NM2TL. For the MLP, we used the same network structure as the MGCN.
    \item For the MVGCN, the network structure and hyperparameter settings were used as the GCN part of our MGCN. The main differences were that MVGCN applied the view pooling \cite{zhang2018multi} to fuse the embedded feature matrices from different modalities without manifold based regularization term.
\end{itemize}
The predictive performance was evaluated using bootstrap analysis on $10$ repeated experiments. We showed the comparison results in Table \ref{compare other models}, from which we can see that our model achieved superior RMSE and MAE over other approaches. Furthermore, enforcing the manifold based regularization term in MGCN improves prediction performance, suggesting that our model can extract more discriminative features by incorporating subjects' relationships within and between different modalities.

\subsection{Model explanation and biomarker identification }
We next interpreted our model and identified the cognition-related biomarkers at the ROI, FC, and FN levels. The following abbreviations of the FNs were used: Sensory/somatomotor network (SMN), Auditory network (AUD), cingulo-opercular task control Network (CNG), default mode network (DMN), visual network (VIS),  Fronto-parietal Task Control (FPN), salience network (SAL), subcortical network (SCT), Ventral attention Network (VTRN), Dorsal attention network (DA), and Cerebellar (CEB).
\subsubsection{ROI identification}
We applied Grad-RAM to identify the ROIs, which played essential roles in WRAT prediction. The Grad-RAM values $\bm{a}^{(m)}$ for 264 ROIs were further normalized by Z-score normalization. Next, we reported the visualization results separately for each fMRI paradigm. Specifically, we considered the top $5\%$ ROIs with the largest Grad-RAM values and further  analyzed the FN segregation with the selected ROIs in Table \ref{ROIs}. In addition, the BrainNetViewer \cite{xia2013brainnet} was used to visualize the ROI identification results in anatomical space, shown in Fig.\ref{anatomyview}.

\subsubsection{FC identification}
Moreover, we interpreted our model by identifying the significant cognition-related FCs using edge mask learning. Because two fMRI paradigms were mapped using the same brain template, we only investigated a single edge mask for both fMRI paradigms. For edge mask learning, the model was retrained using all subjects until the loss function in Eq.\ref{mask loss} converged. We remain all other hyperparameters the same as the previous experiment and set the $L_{1}$ regularization parameters $\beta$ in a grid of $\{0.05, 0.1, 0.5, 1, 5, 10\}$ to control the sparsity of the edge mask. For each value in the grid, we trained the model to learn the mask matrices for 10 times. The edge mask matrices were first binarized that we set the positive elements to 1 and other non-positive entries to 0. Then, we calculated the frequency of the appearance for each FC and showed the binary edge mask with a threshold of $0.5$ in Fig.\ref{mask plot}. We selected $\beta$ based on the sparsity of the edge mask. Consequently, we set $\beta$ $0.5$ and $1$ with the sparsity of the edge mask between $5\%$ and $1\%$, shown in Fig.\ref{mask thre}. Next, we qualitatively analyzed the FN integration by constructing the module allegiance matrices \cite{mattar2015functional} and calculating the numbers of FCs for different $\beta$ within each FN and between each pair of FNs. The pair of FNs with large value in the module allegiance matrix suggests that two FNs may be part of the same function-related community, shown in Fig.\ref{FN map}.

\subsection{Discussion}
\label{Discussion}
In this work, we used a proposed MGCN framework to predict the WRAT scores using both emotion and nback fMRI. We validated our model on the PNC dataset and compared our approach with other methods. The results showed that our MGCN got significantly improved predictive performance over GCN as well as the other competing approaches. Next, we applied Grad-RAM and edge mask learning to interpret our model and identify the significant ROIs and FCs, which played significant roles in predicting WRAT scores. 
Based on the results of ROI identification, we further measured the brain segregation and identified 7 FNs related to human cognitive variability. The cognition-related functions of these FNs were reported by previous studies based on different measures. 
\begin{itemize}
    \item \textbf{SMN}: According to the study in \cite{feng2019verbal, chenji2016investigating}, the recruitment of the bilateral postcentral gyrus from SMN, which is also identified by our MGCN (23 Postcentral L and 32 Postcentral R), plays a significant role in multiple cognitive processes, such as verbal creativity, memory retrieval, imaginative process, cognitive control.
    \item \textbf{CNG}: The CNG network, which is associated with output gating of memory, plays a more downstream role in cognitive control and maintaining cognitive faculties available for current processing requirements\cite{sadaghiani2015functional, wallis2015frontoparietal}.
    \item \textbf{DMN}: Based on the anatomical structure, DMN locates at the top of a cortical hierarchy and is relatively isolated from systems directly involved in perception and action, which suggests it plays an integrative role in cognition \cite{fox2005human,buckner2008brain}. More evidence \cite{sormaz2018default, fox2005human} shows that DMN performs the significant function of navigating social interaction, planning for the future using the experiences.
    \item \textbf{VIS}: During some specific learning processes, there is communication among visual cortex regions in the early sessions when subjects are still getting familiar with the visual cues and do not register high rates of task completion\cite{bogdanov2017learning}. Another possible reason for the identification of the visual network may be the scenario of collecting data \cite{satterthwaite2016philadelphia, satterthwaite2014neuroimaging}. When the fMRI data was collected, the subjects were asked to fixate their eyes on the crosshair, which may involve the recruitment of visual networks. 
    \item \textbf{FPN}: Based on recent studies \cite{zanto2013fronto, wallis2015frontoparietal}, as a flexible hub of cognitive control, FRN can not only reflect the commitment of specific tasks but also serve as a code that can promote learning novel tasks.
    \item \textbf{SAL}: According to the study in\cite{toga2015brain, craig2009you, menon2010saliency}, the SAL can perform multiple complex brain cognitive functions, including self-awareness, emotional, and cognitive information.
    \item \textbf{SCT}: Studies in \cite{koshiyama2018role, wu2019brain} show that the subcortical network has a pivotal role in human cognitive, affective, and social functions. SCT also associates with human language control \cite{bridges2013role}.
\end{itemize}
In addition, we identified the FC functioned most significantly in the cognition-related task using edge mask learning. Results are shown in Fig.\ref{FN map}:
\begin{itemize}
    \item The segregation of distinct networks with stronger within FNs and weaker between FNs connectivity can be observed with a lager $\beta$ value.
    \item When increasing $\beta$ to $10$, only the diagonal element remained in the mask. Thus, the model identified the most significant cognition-related FNs, including SMN, CNG, DMN, VIS, FPN, SAL, SCT, which are identical with the identification results obtained by the Grad-RAM approach. These results cross-validated the effectiveness of both model interpretation approaches.
    \item We observed high integration between FPN and SAL. A strong relationship between these two FNs has been reported in the previous study \cite{seeley2007dissociable},  suggesting that the FPN may be part of the SAL \cite{seeley2007dissociable}. Specifically, FPN and SAL respond to the salient environmental stimuli, stimulating the cascade of cognitive control signals \cite{menon2010saliency} and collaborating for the better task accuracy\cite{elton2014divergent}.
    \item The integration between DMN and FPN can be observed from our results. Studies suggest that the integration between DMN and FPN contributes to the mind-wandering state associating with cognitive control \cite{fox2015wandering, golchert2017individual}.
\end{itemize}

\section{Conclusion}
\label{Conclusion}
In this paper, we proposed an interpretable MGCN model for the joint analysis of multi-modal data. To simultaneously incorporate the relationships of subjects within and between modalities, we enforced a manifold based regularization term. We validated our proposed model on the PNC dataset by using multiple paradigms of fMRI. The experimental results demonstrated the superior prediction performance compared with other competing models, indicating that the MGCN can efficiently learn the graph embeddings from multi-modal data and boost the prediction performance. In addition, we proposed Grad-RAM and edge mask learning to qualitatively interpret our model at different levels and identify the significant cognition-related biomarkers. The results from Grad-RAM and the edge mask learning cross-validated one another. The identified biomarkers are supported by the previous reports and may partially account for the variance in human cognitive function.

\ifCLASSOPTIONcaptionsoff
  \newpage
\fi

\bibliography{IEEEabrv,IEEEexample}

\end{document}